\documentclass[11pt, a4paper, twocolumn]{article}
\usepackage{geometry}
\usepackage[utf8]{inputenc} % allow utf-8 input
\usepackage[T1]{fontenc}    % use 8-bit T1 fonts
\usepackage{hyperref}       % hyperlinks
\usepackage{url}            % simple URL typesetting
\usepackage{booktabs}       % professional-quality tables
\usepackage{amsfonts}       % blackboard math symbols
\usepackage{amsmath}
\usepackage{nicefrac}       % compact symbols for 1/2, etc.
\usepackage{microtype}      % microtypography
\usepackage{apacite}
\usepackage{mathptmx}
\usepackage{multicol}

\usepackage{graphicx} 
\graphicspath{{media/}}

\usepackage{float}
\usepackage{tabularx}
\usepackage[flushleft]{threeparttable}
\usepackage[
singlelinecheck=false % left align captions
]{caption}

\title{Latenrgy: Model Agnostic Latency and Energy Consumption Prediction for Binary Classifiers}

\author{
 Jason M. Pittman \\
  University of Maryland Global Campus\\
  \texttt{https://orcid.org/0000-0002-5198-8157} \\ \\
}
\date{} % empty to remove date from title

\begin{document}
\maketitle

%\begin{multicols}{2}

\textbf{
\textit{Abstract} - Machine learning systems increasingly drive innovation across scientific fields and industry, yet challenges in compute overhead—specifically during inference—limit their scalability and sustainability. Responsible AI guardrails, essential for ensuring fairness, transparency, and privacy, further exacerbate these computational demands. This study addresses critical gaps in the literature, chiefly the lack of generalized predictive techniques for latency and energy consumption, limited cross-comparisons of classifiers, and unquantified impacts of RAI guardrails on inference performance. Using Theory Construction Methodology, this work constructed a model-agnostic theoretical framework for predicting latency and energy consumption in binary classification models during inference. The framework synthesizes classifier characteristics, dataset properties, and RAI guardrails into a unified analytical instrument. Two predictive equations are derived that capture the interplay between these factors while offering generalizability across diverse classifiers. The proposed framework provides foundational insights for designing efficient, responsible ML systems. It enables researchers to benchmark and optimize inference performance and assists practitioners in deploying scalable solutions. Finally, this work establishes a theoretical foundation for balancing computational efficiency with ethical AI principles, paving the way for future empirical validation and broader applications.
}

% keywords can be removed

\section{Introduction}
Machine learning (ML) has become integral to diverse scientific fields and business applications. In genomics, ML helps to decode complex genetic patterns, while in climatology, it improves the predictive accuracy of extreme weather events. Across industries, ML is revolutionizing healthcare through diagnostic support and advancing finance via fraud detection systems.

Despite its widespread success, the field of ML faces persistent challenges. One such challenge is compute overhead or the computational resources consumed during the training and inference phases of ML models. Training involves the extensive energy and processing power required to optimize model parameters across large datasets. Inference, on the other hand, focuses on generating predictions from trained models, where compute overhead is characterized by the interplay between latency (the time required to produce a prediction) and energy consumption (the power expended during inference tasks). High latency or energy consumption can limit the scalability, accessibility, and sustainability of ML systems, especially in resource-constrained environments such as mobile and edge devices \cite{henderson2020towards}.

Adding to these challenges is the growing emphasis on Responsible AI (RAI). RAI is a framework of principles aimed at ensuring AI technologies are ethical, fair, and trustworthy. RAI principles include transparency, accountability, fairness, privacy, and robustness \cite{li2024towards}. To operationalize these principles, technical controls and guardrails are employed. While essential for trustworthy AI deployment, these principles impose additional computational burdens during training and inference. Doing so exacerbates existing issues of latency and energy consumption.

Surprisingly given the importance of RAI, the literature offers limited insights into how guardrails in particular impact compute overhead during inference \cite{elesedy2024lora}. While this gap may seem abstract at a broad level, it becomes highly relevant in specific scenarios, such as binary classification models deployed in resource-sensitive environments. Understanding these impacts is critical for guiding the design and scaling of ML systems in both scientific and industrial contexts.

This study is motivated by three specific challenges within the broader gap. First, there is a lack of generalized predictive techniques for estimating classifier latency and energy consumption \cite{mallik2023epam}. Second, limited cross-comparison of classification algorithms has hindered understanding of how different models contribute to these overheads \cite{cassales2022balancing}. Finally, the potential impacts of RAI guardrails, such as explainability and interpretability mechanisms, on inference latency and energy consumption remain underexplored \cite{li2024towards}.

In response to these challenges, this work sought to construct a model-agnostic equation for predicting latency and energy consumption in binary classification models during inference with RAI guardrails. By addressing these issues, this study contributes a theoretical foundation for optimizing compute overhead while balancing the computational efficiency and ethical robustness of ML systems.

The remainder of this paper is organized as follows: Section \ref{sec:2} reviews related work, providing a foundation of background research. Section \ref{sec:3} details the theoretical methodology used to derive the predictive equation. Section \ref{sec:4} presents the derived equation and its components. Finally, Section \ref{sec:5} concludes with a discussion of the study’s implications and directions for future research.

\section{Related work} \label{sec:2}
A comprehensive understanding of this study’s contribution requires familiarity with three key topics: benchmarking ML compute overhead, the trade-off between latency and energy consumption, as well as the foundation for RAI. The following sections summarize seminal and highly influential works in each topic. Such existing literature provides necessary context and grounding for this study’s theoretical framework and its focus on model-agnostic predictions of compute overhead. 

\subsection{Benchmarking ML Compute Overhead}
Benchmarking compute overhead in machine learning (ML) is essential for understanding and optimizing the performance and efficiency of ML systems across diverse tasks and deployment scenarios. Compute overhead encompasses the computational resources consumed during both training and inference phases, with significant implications for scalability, sustainability, and accessibility \cite{strubell2020energy, henderson2020towards}. While training requires substantial resources to optimize model parameters, inference focuses on generating predictions in real-time. With inference, metrics such as latency (prediction time) and energy consumption (power usage) are critical \cite{mattson2020mlperf, reddi2020mlperf} to total cost of ownership and user experience. Thus, effective benchmarking provides a foundation for evaluating and improving ML systems where achieving low latency and high energy efficiency is paramount \cite{cassales2022balancing, mallik2023epam}. Additionally, benchmarks such as MLPerf and related studies have emphasized the growing importance of quantifying compute overhead to address operational efficiency and environmental impact \cite{tschand2024mlperf}.

A critical distinction exists between compute overhead during training and inference. Training involves iterative optimization over large datasets, requiring substantial computational resources and prolonged processing times \cite{strubell2020energy}. Inference, by contrast, focuses on real-time applications, where latency (the time required to produce a prediction) and energy consumption (the power required to perform inference) are paramount \cite{henderson2020towards}). Although the literature has traditionally emphasized the training phase, inference has received comparatively less attention.

To address some of these challenges, benchmarking frameworks such as MLPerf have been developed. MLPerf provides comprehensive benchmarks for both training and inference, enabling standardized performance evaluations across hardware and software platforms \cite{mattson2020mlperf}. The MLPerf Inference Benchmark evaluates system performance on tasks such as image classification and object detection, offering insights into latency and energy efficiency across different implementations \cite{reddi2020mlperf}. Further, MLPerf Power introduces methodologies for assessing energy efficiency, reflecting the growing concern over the environmental impact of AI workloads \cite{tschand2024mlperf}. While these benchmarks are instrumental in understanding empirical performance, they focus on specific tasks and lack predictive models that generalize across classifiers or operational contexts.

Despite advancements in benchmarking, significant gaps remain. First, current benchmarks such as MLPerf provide empirical performance data but do not offer generalized predictive techniques for estimating latency and energy consumption across classifiers. This limitation hinders the ability to anticipate performance bottlenecks or energy demands in novel deployment scenarios, particularly those involving Responsible AI (RAI) guardrails \cite{mallik2023epam}. Second, no universally accepted metrics exist for comparing latency and energy consumption across ML frameworks and hardware configurations, making cross-platform evaluations inconsistent \cite{mattson2020mlperf}.

Additionally, the literature on benchmarking compute overhead demonstrates a limited cross-comparison of classification algorithms (e.g., SVM, k-Nearest Neighbors, Random Forest, and Neural Networks) concerning their effects on latency and energy consumption. Most studies focus on single-model architectures or narrowly compare a few model types \cite{cassales2022balancing}. This narrow scope restricts generalizability, leaving gaps in understanding how diverse classifiers perform in terms of computational efficiency across real-world scenarios. Addressing these limitations requires a theoretical framework capable of predicting latency and energy consumption in a model-agnostic manner. Doing so also requires an understanding of the inherent tradeoff between latency and energy consumption during inference on ML models.

\subsection{The Latency and Energy Consumption Tradeoff}
The relationship between latency and energy consumption during machine learning inference is complex, often involving trade-offs influenced by model architecture, hardware, and optimization strategies. Generally, reducing latency requires increased computational resources, which can lead to higher energy consumption. Conversely, minimizing energy usage may involve techniques that introduce additional processing time, thereby increasing latency. This inverse relationship is particularly evident in resource-constrained environments, such as edge devices, where balancing performance and efficiency is critical.

Recent studies have explored the trade-off between latency and energy consumption during machine learning inference, with varying levels of generalizability across classification algorithms. For instance, researchers examining multilayer perceptrons (MLPs) demonstrated that hyperparameter optimization could significantly reduce energy consumption during inference with minimal impact on classification accuracy \cite{desislavov2021compute}. By tuning model complexity, such as reducing hidden layers or using lower-precision arithmetic, the study highlights strategies that, while tested on MLPs, may generalize to other model architectures. However, the reliance on specific algorithmic properties limits the immediate applicability of these findings to non-neural network classifiers.

In contrast, Hauschild and Hellbrück \cite{hauschild2022latency} analyzed convolutional neural networks (CNNs) deployed on Internet of Things (IoT) edge devices, emphasizing the dependency of latency and energy consumption on model complexity and wireless data rates. The results show that simplifying CNN architectures can yield substantial efficiency gains in resource-constrained environments, underscoring the importance of tailoring models to deployment scenarios. However, this approach is tightly coupled to CNNs and does not address broader classification paradigms, such as decision trees or support vector machines.

While these studies offer valuable insights into optimizing latency and energy efficiency, the work reflects a broader trend in the literature of focusing on specific models or hardware configurations \cite{cassales2022balancing, tschand2024mlperf}. This limitation underscores the need for generalized predictive techniques that span diverse classification algorithms, bridging the gap between theoretical models and empirical benchmarks. Addressing this challenge is critical for advancing the scalability and efficiency of ML systems, particularly as the integration of Responsible AI (RAI) guardrails introduces additional computational overhead.

\subsection{RAI Controls and Guardrails}
Put simply, RAI ensures AI systems are developed and deployed in ways that are ethical \cite{floridi2018ai4people,mittelstadt2016ethics}. Ethical, in this context, includes fairness, transparency, privacy, security, and trustworthiness as core principles. The idea is an AI system can be considered responsible when the set of relevant principles are present. Here, one should consider \textit{present} as technical continuous monitoring.

To that end, ethical principles have experienced rapid theoretical and practical expansion. In a short time, researchers have developed robust technical frameworks to measure and evaluate these principles. Two prominent examples are the Microsoft Responsible Toolbox and the IBM AI 360 Toolkit. Yet, as much as AI practitioners can use these frameworks to evaluate models, researchers \cite{radclyffe2023assessment, lu2024responsible} suggest RAI is one of the most critical challenges present in AI and ML.

Culturally, the rapid expansion has been motivated by demonstrable harm arising from a lack of RAI. Such examples include discriminatory sentencing and parole decisions in the US justice system \cite{angwin2022machine} as well as Amazon's recruitment tool \cite{dastin2022amazon}. Increasing legal and regulatory requirements such as the US President's Executive Order and the EU’s AI Act \cite{worsdorfer2023eu} are also driving RAI research.

Meanwhile, the literature \cite{khan2022ethics,alzubaidi2023towards} has coalesced around five specific RAI principles: explainability, bias or fairness, robustness or safety, transparency or interpretability, and privacy. Additional principles, such as explicability \cite{prem2023ethical} and accountability \cite{liu2022trustworthy}, have been studied but ultimately fall within the scope of one or more of the five specific principles. Consequently, industry (IBM, Microsoft, US Department of Defense) has settled on explainability, bias, robustness, interpretability, and privacy for practical implementation of RAI. Trustworthiness tends to be discussed as an emergent principle only present when the complete set of RAI principles have sound implementations. 

On that note, the RAI principles can be implemented either as a \textit{control} or \textit{guardrail}. On the one hand, controls are techniques applied during the training phase of a model to ensure that the AI system behaves ethically and responsibly \cite{mitchell2019model, mehrabi2021survey}. On the other hand, guardrails are measures implemented in deployed models to assess the runtime behavior of models \cite{raji2019actionable, varshney2017safety}. The aim is to ensure that the AI system continues to operate responsibly and ethically throughout the life of the system deployment \cite{holstein2019improving}.

Despite the stated need for RAI and the availability of broad technical frameworks, the computational costs of implementing these guardrails are often excluded from benchmarking studies. For example, the additional overhead introduced by explainability mechanisms during inference remains an under explored area \cite{li2024towards}. Without incorporating RAI considerations, existing benchmarks risk becoming outdated or incomplete as the adoption of RAI increases. Moreover, and perhaps most importantly, the field is bereft of operationally validated knowledge of how runtime RAI may be more of a poison than a cure.

\section{Method} \label{sec:3}
This work was motivated by a single research question: What variables, coefficients, and propositional operations are necessary for a model-agnostic equation to be capable of predicting latency and energy consumption in binary classification models during inference with RAI guardrails? To answer this question, the study employed Theory Construction Methodology (TCM) to derive the model-agnostic equation.

TCM is a structured approach to developing theoretical frameworks by defining key variables, establishing relationships, and formalizing them into mathematical models \cite{dubin1978theory}. While TCM has been widely applied in theoretical modeling, its application to derive predictive equations for latency and energy consumption in the context of RAI guardrails represents a novel adaptation of this methodology. This approach is particularly well-suited to the research problem because the abstraction and generalization required for a predictive equation applicable across diverse classifiers necessitates a theoretical framework \cite{kaplan2019siri}.

The TCM process began with identifying core variables influencing latency and energy consumption during inference. These variables were selected based on prior empirical findings and theoretical reasoning, ensuring relevance to diverse classification contexts and computational scenarios. For example, the computational overhead introduced by explainability and interpretability guardrails, such as those implemented using SHAP \cite{lundberg2017unified} or LIME \cite{ribeiro2016should}, was identified as a critical variable. This assumption is supported by computational complexity theory, which posits that even linear increases in input size ($O(n)$) result in proportional growth in computational demand. In the context of RAI guardrails, the overhead arises from explainability mechanisms that augment inference operations with additional interpretive computations.

Relationships among these variables—such as the inverse correlation between latency and energy consumption—are then proposed based on prior research \cite{henderson2020towards, mallik2023epam}. For instance, studies such as those by Hauschild and Hellbrück \cite{hauschild2022latency} demonstrate how computational trade-offs between latency and energy efficiency are particularly evident in edge computing environments. Coefficients are incorporated to represent adjustable factors, including the type of classifier and specific deployment conditions. These variables and coefficients are connected through mathematical operations, such as additive and multiplicative terms, to capture their interactions \cite{cassales2022balancing}.

Finally, the equation is formalized to ensure generalizability, interpretability, and scalability across classifiers such as SVM, k-Nearest Neighbors, Random Forest, and Neural Networks. This theoretical framework establishes a foundation for subsequent empirical validation, where its predictive accuracy will be tested against experimental data in diverse operational settings.

\section{Discussion} \label{sec:4}
The development of a model-agnostic equation for predicting latency and energy consumption began with identifying foundational variables (Table \ref{table:1}). These variables are organized into three sets—classification algorithm, RAI guardrail, and dataset characteristics—all of which serve as inputs to a prediction function $f$. The outputs of the function, latency ($L$) and energy consumption ($E$), are represented collectively as $O$.

\begin{table*}[htbp!]
    \centering
    \begin{threeparttable} \caption{Foundational variables in a model-agnostic equation}
        \begin{tabular}{@{}p{7cm}lc@{}}
            \toprule
            {\textbf{Variable Set}} & \textbf{Symbol} \\
            \hline
            \hline
            Classification algorithm & $A$ \\
            RAI guardrail & $G$ \\
            Dataset characteristics & $D$ \\
            Output metric & $O$ \\
            \bottomrule
            \end{tabular}
            \label{table:1}
        \begin{tablenotes}
        \item \textit{Note:} The prediction function $f$ is undefined in the general equation. The formalized prediction equations for $L$ and $E$ are outlined in Table \ref{table:2}.
        \end{tablenotes}
    \end{threeparttable}
\end{table*}

\subsection{General Equation}
A general equation (1) was constructed to unify the dimensions of latency and energy consumption into a cohesive analytical framework:

\begin{equation}
O=f(A,D,G)
\end{equation}

This equation serves two purposes. First, it provides a unified framework to compare inference performance across binary classifiers. Second, it establishes a foundation for synthesizing disparate dimensions of model performance into a predictive tool, enabling cross-model comparisons, performance prediction, and the integration of RAI guardrails into system design.

\subsection{Expanded Variables}
Each variable set in the general equation is expanded into measurable elements. Algorithm type ($A$) contains four discrete elements: support vector machines (SVM), k-nearest neighbors (k-NN), random forests (RF), and neural networks (NN). Categorical encoding is used to represent binary classifiers as $a \in {SVM, k\text{-}NN, RF, NN}$, with $A$ encoded as ${1,0,0,0}$ to predict $L$ or $E$ for SVM, for instance.

Dataset characteristics ($D$) include the number of samples ($n$), feature dimensionality ($p$), and data type ($t$). Data type is represented as a categorical variable with tabular data encoded as 0, text as 1, and image data as 2.

RAI guardrails ($G$) encompass five principles: explainability, fairness, interpretability, safety, and privacy. Each principle is modeled as a binary state ($[0, 1]$), which, when active, can include a continuous intensity score. For example, explainability ($expl$) could take a value of $0.7$, representing partial feature-level explanations covering the top 70\% of features.

\subsection{Prediction Equations}
The general equation was expanded into two prediction equations, capturing latency ($L$) and energy consumption ($E$). These equations model inference performance as a function of algorithm type, dataset characteristics, and the computational cost of RAI guardrails.

The latency equation (2) incorporates logarithmic scaling for dataset size, capturing the diminishing impact of larger datasets on prediction time:

\begin{equation}
L = \alpha + \beta_A A + \beta_D \log(n) + \gamma_D p + \delta_D t + \sum_i \phi_{G,i} g_i + \epsilon
\end{equation}

The energy consumption equation (3) applies linear scaling for dataset size to account for cumulative resource demands during inference:

\begin{equation}
E = \alpha^\prime + \beta_A A + \beta_D^\prime n + \gamma_D p + \delta_D t + \sum_i \phi^\prime_{G,i} g_i + \epsilon^\prime
\end{equation}

Both equations use coefficients to model the contribution of each variable, as summarized in Table \ref{table:2}.

\begin{table*}[!htbp]
    \centering
    \begin{threeparttable} \caption{Coefficients for model-agnostic prediction equations}
        \begin{tabular}{@{}p{7cm}lcc@{}}
            \toprule
            {\textbf{Coefficient Set}} & \textbf{Symbol} & \textbf{Variable} \\
            \hline
            \hline
            Baseline inference & $\alpha, \alpha^\prime$ & $O$ \\
            Error terms for variability$^1$ & $\epsilon, \epsilon^\prime$ & - \\
            Algorithm type & $\beta_A, \beta_A^\prime$ & $A$ \\
            Dataset size & $\beta_D, \beta_D^\prime$ & $D_n$ \\
            Feature dimensionality & $\gamma_D, \gamma_D^\prime$ & $D_p$ \\
            Dataset type & $\delta_D, \delta_D^\prime$ & $D_t$ \\
            Guardrails & $\phi_{G,i}, \phi_{G,i}^\prime$ & $G$ \\
            \bottomrule
        \end{tabular}
        \label{table:2}
        \begin{tablenotes}
    \item \textit{Note:} $^1$ Error terms handle unmodeled variability during inference.
    \end{tablenotes}
    \end{threeparttable}
\end{table*}

\subsection{Novelty and Practical Implications}

These equations provide a novel approach to predicting inference performance across diverse binary classifiers. Unlike prior studies, which focus on empirical benchmarking or specific algorithms \cite{cassales2022balancing, mallik2023epam}, this framework offers generalizability and scalability. Furthermore, it uniquely integrates the computational cost of RAI guardrails, addressing a critical gap in the literature \cite{li2024towards, ribeiro2016should}.

Future empirical validation will use benchmarks such as MLPerf \cite{mattson2020mlperf} to evaluate the predictive accuracy of these models. Practical applications include optimizing ML systems for edge devices, estimating resource demands for RAI-integrated classifiers, and enabling informed trade-offs between latency, energy consumption, and ethical robustness.

\section{Conclusion} \label{sec:5}
AI broadly, and ML in specific, continues to transform science and industry. Yet, AI and ML scalability and accessibility are often constrained by compute overhead. The literature suggests such issues are particularly notable during inference. Challenges such as the lack of generalized predictive techniques for latency and energy consumption, limited cross-comparison of classification algorithms, and the unquantified computational impact of RAI guardrails have left critical gaps in the literature. This study aimed to address these gaps by developing a model-agnostic equation capable of predicting latency and energy consumption in binary classification models during inference with RAI guardrails.

The key contributions of this work include a model-agnostic theoretical framework for analyzing inference performance and two predictive equations for latency and energy consumption. These models synthesize algorithm characteristics, dataset properties, and the computational overhead of RAI guardrails into a cohesive analytical tool. Unlike previous studies that focus on specific classifiers or empirical benchmarks, this work offers generalizability and scalability, bridging theoretical modeling with practical performance evaluation.

The broader significance of this research lies in its implications for designing and deploying efficient, responsible ML systems. For researchers, the predictive equations provide a foundational tool for benchmarking and optimizing inference performance across diverse classifiers. For practitioners, they enable informed decisions about deploying models in resource-constrained environments, such as edge or mobile devices, while maintaining ethical robustness. This work also aligns with the growing need for sustainable AI, offering a pathway to balance computational efficiency with ethical considerations.

In conclusion, this study provides a theoretical foundation for understanding and predicting inference performance in ML systems. By addressing critical gaps in the literature, it lays the groundwork for future advancements in model-agnostic performance prediction, enabling the next generation of scalable and responsible AI systems.

\subsection{Limitations}
While this study provides a foundational framework for predicting inference latency and energy consumption in binary classification models, five limitations should be acknowledged.  

First, the prediction equations rely on assumptions about variable relationships, such as logarithmic scaling for dataset size in latency prediction and linear scaling for energy consumption. While these assumptions are grounded in prior research and theoretical reasoning, they may not fully capture real-world complexities in all scenarios. Additional research, more especially practical experimentation may reveal to what extent such a limitation is addressable.

Second, the focus on binary classification tasks excludes multi-class classification and other ML tasks, such as regression or clustering, which may involve different computational trade-offs. Along similar thinking, this work does not account for potential innovations becoming available in the future.

Third, the representation of RAI guardrails, while practical, simplifies potential computational impact. Complex guardrails, such as differential privacy or trustworthiness mechanisms, may require more nuanced modeling to fully capture resource demands.

Fourth, the framework abstracts dataset characteristics to size, feature dimensionality, and data type. Other important factors, such as data quality or sparsity, are not included and could affect predictions in specific contexts.

Finally, this study presents theoretical equations without empirical validation. While the models are rigorous, their accuracy and generalizability remain untested. Future work will involve validating these equations with experimental data across diverse classifiers, datasets, and deployment environments to ensure their practical applicability.

\subsection{Future work}
There are several areas for future work based on the theoretical framework demonstrated in this research.

Foremost, experimentation is necessary to validate and quantify the coefficients in the latency ($L$) and energy consumption ($E$) prediction equations. Empirical studies using benchmark datasets and platforms such as MLPerf will help calibrate these coefficients, ensuring their accuracy across diverse classifiers and deployment environments. Validation efforts should also explore the sensitivity of the equations to different input variables, such as dataset characteristics and RAI guardrails, to refine the models further.

Furthermore, the generalizability of the $L$ and $E$ predictive equations may be investigated by varying the set $A$ across a variety of AI subfields. Of particular interest, given the mainstream perception of AI, might be the application of the framework to Large Language Models (LLMs), where inference latency and energy efficiency are critical due to their size and complexity. Additionally, frontier research areas such as neuro-symbolic AI represent a compelling opportunity for extending the framework to hybrid models that combine symbolic reasoning with deep learning. These extensions could provide valuable insights into the computational trade-offs in emerging AI paradigms.

Another avenue for future work involves refining the representation of RAI guardrails. Current binary and intensity-scale representations may oversimplify the computational demands of advanced guardrails, such as differential privacy, adversarial robustness, or nuanced interpretability mechanisms. Developing more granular or context-aware models for guardrail contributions could enhance the framework’s precision and applicability.

Finally, while this study focused on binary classification tasks, future research could extend the framework to multi-class classification and other ML tasks, such as regression or clustering. These extensions would test the framework’s scalability and adaptability, addressing broader applications in AI.

\bibliographystyle{apacite}
\bibliography{references.bib}

\begin{thebibliography}{}

\bibitem [\protect \citeauthoryear {%
Alzubaidi%
\ \protect \BOthers {.}}{%
Alzubaidi%
\ \protect \BOthers {.}}{%
{\protect \APACyear {2023}}%
}]{%
alzubaidi2023towards}
\APACinsertmetastar {%
alzubaidi2023towards}%
\begin{APACrefauthors}%
Alzubaidi, L.%
, Al-Sabaawi, A.%
, Bai, J.%
, Dukhan, A.%
, Alkenani, A\BPBI H.%
, Al-Asadi, A.%
\BDBL {}others%
\end{APACrefauthors}%
\unskip\
\newblock
\APACrefYearMonthDay{2023}{}{}.
\newblock
{\BBOQ}\APACrefatitle {Towards Risk-Free Trustworthy Artificial Intelligence: Significance and Requirements} {Towards risk-free trustworthy artificial intelligence: Significance and requirements}.{\BBCQ}
\newblock
\APACjournalVolNumPages{International Journal of Intelligent Systems}{2023}{1}{4459198}.
\PrintBackRefs{\CurrentBib}

\bibitem [\protect \citeauthoryear {%
Angwin%
, Larson%
, Mattu%
\BCBL {}\ \BBA {} Kirchner%
}{%
Angwin%
\ \protect \BOthers {.}}{%
{\protect \APACyear {2022}}%
}]{%
angwin2022machine}
\APACinsertmetastar {%
angwin2022machine}%
\begin{APACrefauthors}%
Angwin, J.%
, Larson, J.%
, Mattu, S.%
\BCBL {}\ \BBA {} Kirchner, L.%
\end{APACrefauthors}%
\unskip\
\newblock
\APACrefYearMonthDay{2022}{}{}.
\newblock
{\BBOQ}\APACrefatitle {Machine bias} {Machine bias}.{\BBCQ}
\newblock
\BIn{} \APACrefbtitle {Ethics of data and analytics} {Ethics of data and analytics}\ (\BPGS\ 254--264).
\newblock
\APACaddressPublisher{}{Auerbach Publications}.
\PrintBackRefs{\CurrentBib}

\bibitem [\protect \citeauthoryear {%
Cassales%
, Gomes%
, Bifet%
, Pfahringer%
\BCBL {}\ \BBA {} Senger%
}{%
Cassales%
\ \protect \BOthers {.}}{%
{\protect \APACyear {2022}}%
}]{%
cassales2022balancing}
\APACinsertmetastar {%
cassales2022balancing}%
\begin{APACrefauthors}%
Cassales, G.%
, Gomes, H\BPBI M.%
, Bifet, A.%
, Pfahringer, B.%
\BCBL {}\ \BBA {} Senger, H.%
\end{APACrefauthors}%
\unskip\
\newblock
\APACrefYearMonthDay{2022}{}{}.
\newblock
{\BBOQ}\APACrefatitle {Balancing performance and energy consumption of bagging ensembles for the classification of data streams in edge computing} {Balancing performance and energy consumption of bagging ensembles for the classification of data streams in edge computing}.{\BBCQ}
\newblock
\APACjournalVolNumPages{IEEE Transactions on Network and Service Management}{20}{3}{3038--3054}.
\PrintBackRefs{\CurrentBib}

\bibitem [\protect \citeauthoryear {%
Dastin%
}{%
Dastin%
}{%
{\protect \APACyear {2022}}%
}]{%
dastin2022amazon}
\APACinsertmetastar {%
dastin2022amazon}%
\begin{APACrefauthors}%
Dastin, J.%
\end{APACrefauthors}%
\unskip\
\newblock
\APACrefYearMonthDay{2022}{}{}.
\newblock
{\BBOQ}\APACrefatitle {Amazon scraps secret AI recruiting tool that showed bias against women} {Amazon scraps secret ai recruiting tool that showed bias against women}.{\BBCQ}
\newblock
\BIn{} \APACrefbtitle {Ethics of data and analytics} {Ethics of data and analytics}\ (\BPGS\ 296--299).
\newblock
\APACaddressPublisher{}{Auerbach Publications}.
\PrintBackRefs{\CurrentBib}

\bibitem [\protect \citeauthoryear {%
Desislavov%
, Mart{\'\i}nez-Plumed%
\BCBL {}\ \BBA {} Hern{\'a}ndez-Orallo%
}{%
Desislavov%
\ \protect \BOthers {.}}{%
{\protect \APACyear {2021}}%
}]{%
desislavov2021compute}
\APACinsertmetastar {%
desislavov2021compute}%
\begin{APACrefauthors}%
Desislavov, R.%
, Mart{\'\i}nez-Plumed, F.%
\BCBL {}\ \BBA {} Hern{\'a}ndez-Orallo, J.%
\end{APACrefauthors}%
\unskip\
\newblock
\APACrefYearMonthDay{2021}{}{}.
\newblock
{\BBOQ}\APACrefatitle {Compute and energy consumption trends in deep learning inference} {Compute and energy consumption trends in deep learning inference}.{\BBCQ}
\newblock
\APACjournalVolNumPages{arXiv preprint arXiv:2109.05472}{}{}{}.
\PrintBackRefs{\CurrentBib}

\bibitem [\protect \citeauthoryear {%
Dubin%
}{%
Dubin%
}{%
{\protect \APACyear {1978}}%
}]{%
dubin1978theory}
\APACinsertmetastar {%
dubin1978theory}%
\begin{APACrefauthors}%
Dubin, R.%
\end{APACrefauthors}%
\unskip\
\newblock
\APACrefYear{1978}.
\newblock
\APACrefbtitle {Theory Building} {Theory building}.
\newblock
\APACaddressPublisher{}{The Free Press}.
\PrintBackRefs{\CurrentBib}

\bibitem [\protect \citeauthoryear {%
Elesedy%
, Esperan{\c{c}}a%
, Oprea%
\BCBL {}\ \BBA {} Ozay%
}{%
Elesedy%
\ \protect \BOthers {.}}{%
{\protect \APACyear {2024}}%
}]{%
elesedy2024lora}
\APACinsertmetastar {%
elesedy2024lora}%
\begin{APACrefauthors}%
Elesedy, H.%
, Esperan{\c{c}}a, P\BPBI M.%
, Oprea, S\BPBI V.%
\BCBL {}\ \BBA {} Ozay, M.%
\end{APACrefauthors}%
\unskip\
\newblock
\APACrefYearMonthDay{2024}{}{}.
\newblock
{\BBOQ}\APACrefatitle {Lora-guard: Parameter-efficient guardrail adaptation for content moderation of large language models} {Lora-guard: Parameter-efficient guardrail adaptation for content moderation of large language models}.{\BBCQ}
\newblock
\APACjournalVolNumPages{arXiv preprint arXiv:2407.02987}{}{}{}.
\PrintBackRefs{\CurrentBib}

\bibitem [\protect \citeauthoryear {%
Floridi%
\ \protect \BOthers {.}}{%
Floridi%
\ \protect \BOthers {.}}{%
{\protect \APACyear {2018}}%
}]{%
floridi2018ai4people}
\APACinsertmetastar {%
floridi2018ai4people}%
\begin{APACrefauthors}%
Floridi, L.%
, Cowls, J.%
, Beltrametti, M.%
, Chatila, R.%
, Chazerand, P.%
, Dignum, V.%
\BDBL {}others%
\end{APACrefauthors}%
\unskip\
\newblock
\APACrefYearMonthDay{2018}{}{}.
\newblock
{\BBOQ}\APACrefatitle {AI4People—an ethical framework for a good AI society: opportunities, risks, principles, and recommendations} {Ai4people—an ethical framework for a good ai society: opportunities, risks, principles, and recommendations}.{\BBCQ}
\newblock
\APACjournalVolNumPages{Minds and machines}{28}{}{689--707}.
\PrintBackRefs{\CurrentBib}

\bibitem [\protect \citeauthoryear {%
Hauschild%
\ \BBA {} Hellbr{\"u}ck%
}{%
Hauschild%
\ \BBA {} Hellbr{\"u}ck%
}{%
{\protect \APACyear {2022}}%
}]{%
hauschild2022latency}
\APACinsertmetastar {%
hauschild2022latency}%
\begin{APACrefauthors}%
Hauschild, S.%
\BCBT {}\ \BBA {} Hellbr{\"u}ck, H.%
\end{APACrefauthors}%
\unskip\
\newblock
\APACrefYearMonthDay{2022}{}{}.
\newblock
{\BBOQ}\APACrefatitle {Latency and Energy Consumption of Convolutional Neural Network Models from IoT Edge Perspective} {Latency and energy consumption of convolutional neural network models from iot edge perspective}.{\BBCQ}
\newblock
\BIn{} \APACrefbtitle {Global IoT Summit} {Global iot summit}\ (\BPGS\ 385--396).
\newblock
\APACaddressPublisher{}{Springer}.
\PrintBackRefs{\CurrentBib}

\bibitem [\protect \citeauthoryear {%
Henderson%
\ \protect \BOthers {.}}{%
Henderson%
\ \protect \BOthers {.}}{%
{\protect \APACyear {2020}}%
}]{%
henderson2020towards}
\APACinsertmetastar {%
henderson2020towards}%
\begin{APACrefauthors}%
Henderson, P.%
, Hu, J.%
, Romoff, J.%
, Brunskill, E.%
, Jurafsky, D.%
\BCBL {}\ \BBA {} Pineau, J.%
\end{APACrefauthors}%
\unskip\
\newblock
\APACrefYearMonthDay{2020}{}{}.
\newblock
{\BBOQ}\APACrefatitle {Towards the systematic reporting of the energy and carbon footprints of machine learning} {Towards the systematic reporting of the energy and carbon footprints of machine learning}.{\BBCQ}
\newblock
\APACjournalVolNumPages{Journal of Machine Learning Research}{21}{248}{1--43}.
\PrintBackRefs{\CurrentBib}

\bibitem [\protect \citeauthoryear {%
Holstein%
, Wortman~Vaughan%
, Daum{\'e}~III%
, Dudik%
\BCBL {}\ \BBA {} Wallach%
}{%
Holstein%
\ \protect \BOthers {.}}{%
{\protect \APACyear {2019}}%
}]{%
holstein2019improving}
\APACinsertmetastar {%
holstein2019improving}%
\begin{APACrefauthors}%
Holstein, K.%
, Wortman~Vaughan, J.%
, Daum{\'e}~III, H.%
, Dudik, M.%
\BCBL {}\ \BBA {} Wallach, H.%
\end{APACrefauthors}%
\unskip\
\newblock
\APACrefYearMonthDay{2019}{}{}.
\newblock
{\BBOQ}\APACrefatitle {Improving fairness in machine learning systems: What do industry practitioners need?} {Improving fairness in machine learning systems: What do industry practitioners need?}{\BBCQ}
\newblock
\BIn{} \APACrefbtitle {Proceedings of the 2019 CHI conference on human factors in computing systems} {Proceedings of the 2019 chi conference on human factors in computing systems}\ (\BPGS\ 1--16).
\PrintBackRefs{\CurrentBib}

\bibitem [\protect \citeauthoryear {%
Kaplan%
\ \BBA {} Haenlein%
}{%
Kaplan%
\ \BBA {} Haenlein%
}{%
{\protect \APACyear {2019}}%
}]{%
kaplan2019siri}
\APACinsertmetastar {%
kaplan2019siri}%
\begin{APACrefauthors}%
Kaplan, A.%
\BCBT {}\ \BBA {} Haenlein, M.%
\end{APACrefauthors}%
\unskip\
\newblock
\APACrefYearMonthDay{2019}{}{}.
\newblock
{\BBOQ}\APACrefatitle {Siri, Siri, in my hand: Who’s the fairest in the land? On the interpretations, illustrations, and implications of artificial intelligence} {Siri, siri, in my hand: Who’s the fairest in the land? on the interpretations, illustrations, and implications of artificial intelligence}.{\BBCQ}
\newblock
\APACjournalVolNumPages{Business horizons}{62}{1}{15--25}.
\PrintBackRefs{\CurrentBib}

\bibitem [\protect \citeauthoryear {%
Khan%
\ \protect \BOthers {.}}{%
Khan%
\ \protect \BOthers {.}}{%
{\protect \APACyear {2022}}%
}]{%
khan2022ethics}
\APACinsertmetastar {%
khan2022ethics}%
\begin{APACrefauthors}%
Khan, A\BPBI A.%
, Badshah, S.%
, Liang, P.%
, Waseem, M.%
, Khan, B.%
, Ahmad, A.%
\BDBL {}Akbar, M\BPBI A.%
\end{APACrefauthors}%
\unskip\
\newblock
\APACrefYearMonthDay{2022}{}{}.
\newblock
{\BBOQ}\APACrefatitle {Ethics of AI: A systematic literature review of principles and challenges} {Ethics of ai: A systematic literature review of principles and challenges}.{\BBCQ}
\newblock
\BIn{} \APACrefbtitle {Proceedings of the 26th International Conference on Evaluation and Assessment in Software Engineering} {Proceedings of the 26th international conference on evaluation and assessment in software engineering}\ (\BPGS\ 383--392).
\PrintBackRefs{\CurrentBib}

\bibitem [\protect \citeauthoryear {%
Li%
, Liu%
, Yang%
\BCBL {}\ \BBA {} Ren%
}{%
Li%
\ \protect \BOthers {.}}{%
{\protect \APACyear {2024}}%
}]{%
li2024towards}
\APACinsertmetastar {%
li2024towards}%
\begin{APACrefauthors}%
Li, P.%
, Liu, Y.%
, Yang, J.%
\BCBL {}\ \BBA {} Ren, S.%
\end{APACrefauthors}%
\unskip\
\newblock
\APACrefYearMonthDay{2024}{}{}.
\newblock
{\BBOQ}\APACrefatitle {Towards Socially and Environmentally Responsible AI} {Towards socially and environmentally responsible ai}.{\BBCQ}
\newblock
\APACjournalVolNumPages{arXiv preprint arXiv:2407.05176}{}{}{}.
\PrintBackRefs{\CurrentBib}

\bibitem [\protect \citeauthoryear {%
Liu%
\ \protect \BOthers {.}}{%
Liu%
\ \protect \BOthers {.}}{%
{\protect \APACyear {2022}}%
}]{%
liu2022trustworthy}
\APACinsertmetastar {%
liu2022trustworthy}%
\begin{APACrefauthors}%
Liu, H.%
, Wang, Y.%
, Fan, W.%
, Liu, X.%
, Li, Y.%
, Jain, S.%
\BDBL {}Tang, J.%
\end{APACrefauthors}%
\unskip\
\newblock
\APACrefYearMonthDay{2022}{}{}.
\newblock
{\BBOQ}\APACrefatitle {Trustworthy ai: A computational perspective} {Trustworthy ai: A computational perspective}.{\BBCQ}
\newblock
\APACjournalVolNumPages{ACM Transactions on Intelligent Systems and Technology}{14}{1}{1--59}.
\PrintBackRefs{\CurrentBib}

\bibitem [\protect \citeauthoryear {%
Lu%
\ \protect \BOthers {.}}{%
Lu%
\ \protect \BOthers {.}}{%
{\protect \APACyear {2024}}%
}]{%
lu2024responsible}
\APACinsertmetastar {%
lu2024responsible}%
\begin{APACrefauthors}%
Lu, Q.%
, Zhu, L.%
, Xu, X.%
, Whittle, J.%
, Zowghi, D.%
\BCBL {}\ \BBA {} Jacquet, A.%
\end{APACrefauthors}%
\unskip\
\newblock
\APACrefYearMonthDay{2024}{}{}.
\newblock
{\BBOQ}\APACrefatitle {Responsible AI pattern catalogue: A collection of best practices for AI governance and engineering} {Responsible ai pattern catalogue: A collection of best practices for ai governance and engineering}.{\BBCQ}
\newblock
\APACjournalVolNumPages{ACM Computing Surveys}{56}{7}{1--35}.
\PrintBackRefs{\CurrentBib}

\bibitem [\protect \citeauthoryear {%
Lundberg%
}{%
Lundberg%
}{%
{\protect \APACyear {2017}}%
}]{%
lundberg2017unified}
\APACinsertmetastar {%
lundberg2017unified}%
\begin{APACrefauthors}%
Lundberg, S.%
\end{APACrefauthors}%
\unskip\
\newblock
\APACrefYearMonthDay{2017}{}{}.
\newblock
{\BBOQ}\APACrefatitle {A unified approach to interpreting model predictions} {A unified approach to interpreting model predictions}.{\BBCQ}
\newblock
\APACjournalVolNumPages{arXiv preprint arXiv:1705.07874}{}{}{}.
\PrintBackRefs{\CurrentBib}

\bibitem [\protect \citeauthoryear {%
Mallik%
, Wang%
, Xie%
, Chen%
\BCBL {}\ \BBA {} Han%
}{%
Mallik%
\ \protect \BOthers {.}}{%
{\protect \APACyear {2023}}%
}]{%
mallik2023epam}
\APACinsertmetastar {%
mallik2023epam}%
\begin{APACrefauthors}%
Mallik, A.%
, Wang, H.%
, Xie, J.%
, Chen, D.%
\BCBL {}\ \BBA {} Han, K.%
\end{APACrefauthors}%
\unskip\
\newblock
\APACrefYearMonthDay{2023}{}{}.
\newblock
{\BBOQ}\APACrefatitle {EPAM: A predictive energy model for mobile AI} {Epam: A predictive energy model for mobile ai}.{\BBCQ}
\newblock
\BIn{} \APACrefbtitle {ICC 2023-IEEE International Conference on Communications} {Icc 2023-ieee international conference on communications}\ (\BPGS\ 954--959).
\PrintBackRefs{\CurrentBib}

\bibitem [\protect \citeauthoryear {%
Mattson%
\ \protect \BOthers {.}}{%
Mattson%
\ \protect \BOthers {.}}{%
{\protect \APACyear {2020}}%
}]{%
mattson2020mlperf}
\APACinsertmetastar {%
mattson2020mlperf}%
\begin{APACrefauthors}%
Mattson, P.%
, Reddi, V\BPBI J.%
, Cheng, C.%
, Coleman, C.%
, Diamos, G.%
, Kanter, D.%
\BDBL {}others%
\end{APACrefauthors}%
\unskip\
\newblock
\APACrefYearMonthDay{2020}{}{}.
\newblock
{\BBOQ}\APACrefatitle {MLPerf: An industry standard benchmark suite for machine learning performance} {Mlperf: An industry standard benchmark suite for machine learning performance}.{\BBCQ}
\newblock
\APACjournalVolNumPages{IEEE Micro}{40}{2}{8--16}.
\PrintBackRefs{\CurrentBib}

\bibitem [\protect \citeauthoryear {%
Mehrabi%
, Morstatter%
, Saxena%
, Lerman%
\BCBL {}\ \BBA {} Galstyan%
}{%
Mehrabi%
\ \protect \BOthers {.}}{%
{\protect \APACyear {2021}}%
}]{%
mehrabi2021survey}
\APACinsertmetastar {%
mehrabi2021survey}%
\begin{APACrefauthors}%
Mehrabi, N.%
, Morstatter, F.%
, Saxena, N.%
, Lerman, K.%
\BCBL {}\ \BBA {} Galstyan, A.%
\end{APACrefauthors}%
\unskip\
\newblock
\APACrefYearMonthDay{2021}{}{}.
\newblock
{\BBOQ}\APACrefatitle {A survey on bias and fairness in machine learning} {A survey on bias and fairness in machine learning}.{\BBCQ}
\newblock
\APACjournalVolNumPages{ACM computing surveys (CSUR)}{54}{6}{1--35}.
\PrintBackRefs{\CurrentBib}

\bibitem [\protect \citeauthoryear {%
Mitchell%
\ \protect \BOthers {.}}{%
Mitchell%
\ \protect \BOthers {.}}{%
{\protect \APACyear {2019}}%
}]{%
mitchell2019model}
\APACinsertmetastar {%
mitchell2019model}%
\begin{APACrefauthors}%
Mitchell, M.%
, Wu, S.%
, Zaldivar, A.%
, Barnes, P.%
, Vasserman, L.%
, Hutchinson, B.%
\BDBL {}Gebru, T.%
\end{APACrefauthors}%
\unskip\
\newblock
\APACrefYearMonthDay{2019}{}{}.
\newblock
{\BBOQ}\APACrefatitle {Model cards for model reporting} {Model cards for model reporting}.{\BBCQ}
\newblock
\BIn{} \APACrefbtitle {Proceedings of the conference on fairness, accountability, and transparency} {Proceedings of the conference on fairness, accountability, and transparency}\ (\BPGS\ 220--229).
\PrintBackRefs{\CurrentBib}

\bibitem [\protect \citeauthoryear {%
Mittelstadt%
, Allo%
, Taddeo%
, Wachter%
\BCBL {}\ \BBA {} Floridi%
}{%
Mittelstadt%
\ \protect \BOthers {.}}{%
{\protect \APACyear {2016}}%
}]{%
mittelstadt2016ethics}
\APACinsertmetastar {%
mittelstadt2016ethics}%
\begin{APACrefauthors}%
Mittelstadt, B\BPBI D.%
, Allo, P.%
, Taddeo, M.%
, Wachter, S.%
\BCBL {}\ \BBA {} Floridi, L.%
\end{APACrefauthors}%
\unskip\
\newblock
\APACrefYearMonthDay{2016}{}{}.
\newblock
{\BBOQ}\APACrefatitle {The ethics of algorithms: Mapping the debate} {The ethics of algorithms: Mapping the debate}.{\BBCQ}
\newblock
\APACjournalVolNumPages{Big Data \& Society}{3}{2}{2053951716679679}.
\PrintBackRefs{\CurrentBib}

\bibitem [\protect \citeauthoryear {%
Prem%
}{%
Prem%
}{%
{\protect \APACyear {2023}}%
}]{%
prem2023ethical}
\APACinsertmetastar {%
prem2023ethical}%
\begin{APACrefauthors}%
Prem, E.%
\end{APACrefauthors}%
\unskip\
\newblock
\APACrefYearMonthDay{2023}{}{}.
\newblock
{\BBOQ}\APACrefatitle {From ethical AI frameworks to tools: a review of approaches} {From ethical ai frameworks to tools: a review of approaches}.{\BBCQ}
\newblock
\APACjournalVolNumPages{AI and Ethics}{3}{3}{699--716}.
\PrintBackRefs{\CurrentBib}

\bibitem [\protect \citeauthoryear {%
Radclyffe%
, Ribeiro%
\BCBL {}\ \BBA {} Wortham%
}{%
Radclyffe%
\ \protect \BOthers {.}}{%
{\protect \APACyear {2023}}%
}]{%
radclyffe2023assessment}
\APACinsertmetastar {%
radclyffe2023assessment}%
\begin{APACrefauthors}%
Radclyffe, C.%
, Ribeiro, M.%
\BCBL {}\ \BBA {} Wortham, R\BPBI H.%
\end{APACrefauthors}%
\unskip\
\newblock
\APACrefYearMonthDay{2023}{}{}.
\newblock
{\BBOQ}\APACrefatitle {The assessment list for trustworthy artificial intelligence: A review and recommendations} {The assessment list for trustworthy artificial intelligence: A review and recommendations}.{\BBCQ}
\newblock
\APACjournalVolNumPages{Frontiers in artificial intelligence}{6}{}{1020592}.
\PrintBackRefs{\CurrentBib}

\bibitem [\protect \citeauthoryear {%
Raji%
\ \BBA {} Buolamwini%
}{%
Raji%
\ \BBA {} Buolamwini%
}{%
{\protect \APACyear {2019}}%
}]{%
raji2019actionable}
\APACinsertmetastar {%
raji2019actionable}%
\begin{APACrefauthors}%
Raji, I\BPBI D.%
\BCBT {}\ \BBA {} Buolamwini, J.%
\end{APACrefauthors}%
\unskip\
\newblock
\APACrefYearMonthDay{2019}{}{}.
\newblock
{\BBOQ}\APACrefatitle {Actionable auditing: Investigating the impact of publicly naming biased performance results of commercial ai products} {Actionable auditing: Investigating the impact of publicly naming biased performance results of commercial ai products}.{\BBCQ}
\newblock
\BIn{} \APACrefbtitle {Proceedings of the 2019 AAAI/ACM Conference on AI, Ethics, and Society} {Proceedings of the 2019 aaai/acm conference on ai, ethics, and society}\ (\BPGS\ 429--435).
\PrintBackRefs{\CurrentBib}

\bibitem [\protect \citeauthoryear {%
Reddi%
\ \protect \BOthers {.}}{%
Reddi%
\ \protect \BOthers {.}}{%
{\protect \APACyear {2020}}%
}]{%
reddi2020mlperf}
\APACinsertmetastar {%
reddi2020mlperf}%
\begin{APACrefauthors}%
Reddi, V\BPBI J.%
, Cheng, C.%
, Kanter, D.%
, Mattson, P.%
, Schmuelling, G.%
, Wu, C\BHBI J.%
\BDBL {}others%
\end{APACrefauthors}%
\unskip\
\newblock
\APACrefYearMonthDay{2020}{}{}.
\newblock
{\BBOQ}\APACrefatitle {Mlperf inference benchmark} {Mlperf inference benchmark}.{\BBCQ}
\newblock
\BIn{} \APACrefbtitle {2020 ACM/IEEE 47th Annual International Symposium on Computer Architecture (ISCA)} {2020 acm/ieee 47th annual international symposium on computer architecture (isca)}\ (\BPGS\ 446--459).
\PrintBackRefs{\CurrentBib}

\bibitem [\protect \citeauthoryear {%
Ribeiro%
, Singh%
\BCBL {}\ \BBA {} Guestrin%
}{%
Ribeiro%
\ \protect \BOthers {.}}{%
{\protect \APACyear {2016}}%
}]{%
ribeiro2016should}
\APACinsertmetastar {%
ribeiro2016should}%
\begin{APACrefauthors}%
Ribeiro, M\BPBI T.%
, Singh, S.%
\BCBL {}\ \BBA {} Guestrin, C.%
\end{APACrefauthors}%
\unskip\
\newblock
\APACrefYearMonthDay{2016}{}{}.
\newblock
{\BBOQ}\APACrefatitle {" Why should i trust you?" Explaining the predictions of any classifier} {" why should i trust you?" explaining the predictions of any classifier}.{\BBCQ}
\newblock
\BIn{} \APACrefbtitle {Proceedings of the 22nd ACM SIGKDD international conference on knowledge discovery and data mining} {Proceedings of the 22nd acm sigkdd international conference on knowledge discovery and data mining}\ (\BPGS\ 1135--1144).
\PrintBackRefs{\CurrentBib}

\bibitem [\protect \citeauthoryear {%
Strubell%
, Ganesh%
\BCBL {}\ \BBA {} McCallum%
}{%
Strubell%
\ \protect \BOthers {.}}{%
{\protect \APACyear {2020}}%
}]{%
strubell2020energy}
\APACinsertmetastar {%
strubell2020energy}%
\begin{APACrefauthors}%
Strubell, E.%
, Ganesh, A.%
\BCBL {}\ \BBA {} McCallum, A.%
\end{APACrefauthors}%
\unskip\
\newblock
\APACrefYearMonthDay{2020}{}{}.
\newblock
{\BBOQ}\APACrefatitle {Energy and policy considerations for modern deep learning research} {Energy and policy considerations for modern deep learning research}.{\BBCQ}
\newblock
\BIn{} \APACrefbtitle {Proceedings of the AAAI conference on artificial intelligence} {Proceedings of the aaai conference on artificial intelligence}\ (\BVOL~34, \BPGS\ 13693--13696).
\PrintBackRefs{\CurrentBib}

\bibitem [\protect \citeauthoryear {%
Tschand%
\ \protect \BOthers {.}}{%
Tschand%
\ \protect \BOthers {.}}{%
{\protect \APACyear {2024}}%
}]{%
tschand2024mlperf}
\APACinsertmetastar {%
tschand2024mlperf}%
\begin{APACrefauthors}%
Tschand, A.%
, Rajan, A\BPBI T\BPBI R.%
, Idgunji, S.%
, Ghosh, A.%
, Holleman, J.%
, Kiraly, C.%
\BDBL {}others%
\end{APACrefauthors}%
\unskip\
\newblock
\APACrefYearMonthDay{2024}{}{}.
\newblock
{\BBOQ}\APACrefatitle {MLPerf Power: Benchmarking the Energy Efficiency of Machine Learning Systems from Microwatts to Megawatts for Sustainable AI} {Mlperf power: Benchmarking the energy efficiency of machine learning systems from microwatts to megawatts for sustainable ai}.{\BBCQ}
\newblock
\APACjournalVolNumPages{arXiv preprint arXiv:2410.12032}{}{}{}.
\PrintBackRefs{\CurrentBib}

\bibitem [\protect \citeauthoryear {%
Varshney%
\ \BBA {} Alemzadeh%
}{%
Varshney%
\ \BBA {} Alemzadeh%
}{%
{\protect \APACyear {2017}}%
}]{%
varshney2017safety}
\APACinsertmetastar {%
varshney2017safety}%
\begin{APACrefauthors}%
Varshney, K\BPBI R.%
\BCBT {}\ \BBA {} Alemzadeh, H.%
\end{APACrefauthors}%
\unskip\
\newblock
\APACrefYearMonthDay{2017}{}{}.
\newblock
{\BBOQ}\APACrefatitle {On the safety of machine learning: Cyber-physical systems, decision sciences, and data products} {On the safety of machine learning: Cyber-physical systems, decision sciences, and data products}.{\BBCQ}
\newblock
\APACjournalVolNumPages{Big data}{5}{3}{246--255}.
\PrintBackRefs{\CurrentBib}

\bibitem [\protect \citeauthoryear {%
W{\"o}rsd{\"o}rfer%
}{%
W{\"o}rsd{\"o}rfer%
}{%
{\protect \APACyear {2023}}%
}]{%
worsdorfer2023eu}
\APACinsertmetastar {%
worsdorfer2023eu}%
\begin{APACrefauthors}%
W{\"o}rsd{\"o}rfer, M.%
\end{APACrefauthors}%
\unskip\
\newblock
\APACrefYearMonthDay{2023}{}{}.
\newblock
{\BBOQ}\APACrefatitle {The EU’s artificial intelligence act: an ordoliberal assessment} {The eu’s artificial intelligence act: an ordoliberal assessment}.{\BBCQ}
\newblock
\APACjournalVolNumPages{AI and Ethics}{}{}{1--16}.
\PrintBackRefs{\CurrentBib}

\end{thebibliography}

%\end{multicols}

\end{document}